\documentclass{article}

\usepackage{times}
\usepackage{graphicx} 
\usepackage{subfigure} 

\usepackage{natbib}

\usepackage{algorithm}
\usepackage{algorithmic}

\usepackage[table,xcdraw,dvipsnames]{xcolor}
\usepackage[colorlinks]{hyperref}

\usepackage{url}
\usepackage{todonotes}
\usepackage{amsbsy}
\usepackage{amsmath}
\usepackage{amsfonts}
\usepackage{amsmath}%
\usepackage{MnSymbol}%
\usepackage{wasysym}%

\usepackage[cal=dutchcal,
calscaled=1,
bb=boondox,
scr=euler]{mathalfa}

\usepackage{tikz}
\newcommand*\circled[1]{\tikz[baseline=(char.base)]{
            \node[shape=circle,draw,inner sep=2pt] (char) {#1};}}

\usepackage[framemethod=TikZ]{mdframed}
\usepackage{wrapfig}

\usepackage[capitalise]{cleveref}
\crefformat{equation}{(#2#1#3)}

\newcommand{\ie}[0]{\emph{i.e.},~}
\newcommand{\eg}[0]{\emph{e.g.},~}

\newcommand{\defeq}{\ensuremath{\doteq}}

\newcommand{\gr}[1]{\ensuremath{{\mathcal{#1}}}}
\newcommand{\settype}[1]{\ensuremath{\mathbb{#1}}}
\renewcommand{\gg}[0]{\ensuremath{\gr{g}}}
\renewcommand{\aa}[0]{\ensuremath{\gr{a}}}
\newcommand{\hh}[0]{\ensuremath{\gr{h}}}
\newcommand{\GG}[0]{\ensuremath{\gr{G}}}

\newcommand{\HH}[0]{\ensuremath{\gr{H}}}

\newcommand{\aut}[0]{\ensuremath{\mathcal{Aut}}}
\newcommand{\Rel}[0]{\ensuremath{\Delta}}
\newcommand{\Str}[0]{\ensuremath{{\Omega}}}

\newcommand{\xx}[0]{\ensuremath{\mathbf{x}}}
\newcommand{\phis}[0]{\ensuremath{\boldsymbol{\phi}}}
\newcommand{\sigmas}[0]{\ensuremath{\boldsymbol{\sigma}}}
\newcommand{\nn}[0]{\ensuremath{[1,\ldots,{N}]}}

\newcommand{\YY}[0]{\ensuremath{\settype{Y}}}
\newcommand{\XX}[0]{\ensuremath{\settype{X}}}

\newcommand{\EE}[0]{\ensuremath{\settype{E}}}
\newcommand{\NN}[0]{\ensuremath{\{1,\ldots,N\}}}
\newcommand{\Nset}[0]{\ensuremath{\settype{N}}}

\newcommand{\Mset}[0]{\ensuremath{\settype{M}}}
\newcommand{\Nvec}[0]{\ensuremath{\vec{\Nset}}}

\newcommand{\WW}[0]{\ensuremath{\mathbf{W}}}
\newcommand{\ww}[0]{\ensuremath{\mathbf{w}}}

\renewcommand{\theta}[0]{\ensuremath{{w}}}
\newcommand{\pis}[0]{\ensuremath{{\pi}}}
\newcommand{\thetas}[0]{\ensuremath{\mathbf{w}}}
\renewcommand{\Re}[0]{\settype{R}}

\newcommand{\ttt}[1]{\ensuremath{_{#1}}}
\renewcommand{\vec}[1]{\ensuremath{\overrightarrow{#1}}}

\usepackage[thmmarks, amsmath, thref]{ntheorem}
\theoremstyle{plain}
\newtheorem{theorem}{Theorem}[section]

\newtheorem{proposition}[theorem]{Proposition}
\newtheorem{observation}[theorem]{Observation}
\newtheorem{claim}[theorem]{Claim}
\newtheorem{corollary}[theorem]{Corollary}
\theoremstyle{definition}
\newtheorem{definition}{Definition}[section]
\theorembodyfont{\normalfont}
\newtheorem{example}{Example}[section]
\theoremsymbol{\ensuremath{\blacksquare}}
\newtheorem*{proof}{Proof}

\usepackage{mathtools}

\usepackage[most]{tcolorbox}
\newcounter{testexamplectr}
\usepackage{xparse}
\def\exampletext{Example } 

\NewDocumentEnvironment{testexample}{ O{} }
{
  \colorlet{colexam}{red!10!black} 
  \refstepcounter{testexamplectr}
    \newtcolorbox{testexamplebox}{%
    empty,
    title={\exampletext \thetestexamplectr.  #1},
    attach boxed title to top left,
       minipage boxed title,
    boxed title style={empty,size=minimal,toprule=0pt,top=4pt,left=3mm,overlay={}},
    coltitle=colexam,fonttitle=\bfseries,
    before=\par\medskip\noindent,parbox=false,boxsep=0pt,left=3mm,right=0mm,top=2pt,breakable,pad at break=0mm,
       before upper=\csname @totalleftmargin\endcsname0pt, 
    overlay unbroken={\draw[colexam,line width=.5pt] ([xshift=-0pt]title.north west) -- ([xshift=-0pt]frame.south west); },
    overlay first={\draw[colexam,line width=.5pt] ([xshift=-0pt]title.north west) -- ([xshift=-0pt]frame.south west); },
    overlay middle={\draw[colexam,line width=.5pt] ([xshift=-0pt]frame.north west) -- ([xshift=-0pt]frame.south west); },
    overlay last={\draw[colexam,line width=.5pt] ([xshift=-0pt]frame.north west) -- ([xshift=-0pt]frame.south west); },%
    }
\begin{testexamplebox}}
{\end{testexamplebox}\endlist}
\numberwithin{testexamplectr}{section}

\usepackage[framemethod=TikZ]{mdframed}

\mdfdefinestyle{MyFrame}{%
    linecolor=Black,
    outerlinewidth=.3pt,
    roundcorner=5pt,
    innertopmargin=\baselineskip,
    innerbottommargin=\baselineskip,
    innerrightmargin=5pt,
    innerleftmargin=5pt,
    backgroundcolor=RoyalBlue!0!white}

\mdfdefinestyle{MyFrame2}{%
    linecolor=White,
    outerlinewidth=.3pt,
    roundcorner=5pt,
    innertopmargin=\baselineskip,
    innerbottommargin=\baselineskip,
    innerrightmargin=5pt,
    innerleftmargin=5pt,
    backgroundcolor=Gray!10!white}

\usepackage[accepted]{icml2017}

\icmltitlerunning{Equivariance Through Parameter-Sharing}

\begin{document} 

\twocolumn[
\icmltitle{Equivariance Through Parameter-Sharing}

\begin{icmlauthorlist}
\icmlauthor{Siamak Ravanbakhsh}{to}
\icmlauthor{Jeff Schneider}{to}
\icmlauthor{Barnab{\'{a}}s P{\'{o}}czos}{to}
\end{icmlauthorlist}

\icmlaffiliation{to}{School of Computer Science, Carnegie Mellon University, 5000 Forbes Ave., Pittsburgh, PA, USA 15217}
\icmlcorrespondingauthor{Siamak Ravanbakhsh}{mravanba@cs.cmu.edu}


\icmlkeywords{equivariance, parameter-sharing, deep learning, neural networks}

\vskip 0.3in
]

\printAffiliationsAndNotice{} 

\begin{abstract}
We propose to study equivariance in deep neural networks through parameter symmetries.
In particular, given a group $\GG$ that acts discretely on the input and output of a standard neural network layer $\phis_{\WW}: \Re^{M} \to \Re^{N}$,
we show that $\phis_{\WW}$ is equivariant with respect to $\GG$-action iff $\GG$ explains the symmetries of the network parameters $\WW$.
Inspired by this observation, we then propose two parameter-sharing schemes to induce the desirable symmetry on $\WW$.
Our procedure for tying the parameters achieves $\GG$-equivariance and, under some conditions on the action of $\GG$, it guarantees sensitivity 
to all other permutation groups outside $\GG$.
\end{abstract}

Given enough training data, a multi-layer perceptron would eventually learn the domain invariances
in a classification task. Nevertheless, success of convolutional and recurrent networks
suggests that encoding the domain symmetries through shared parameters can significantly boost
the generalization of deep neural networks. 
The same observation can be made in deep learning for semi-supervised and
unsupervised learning in structured domains.
This raises an important question that is addressed in this paper:
\textbf{What kind of priors on input/output structure can be encoded through parameter-sharing?}

\begin{figure*}[ht]
  \centering
  \includegraphics[width=.98\linewidth]{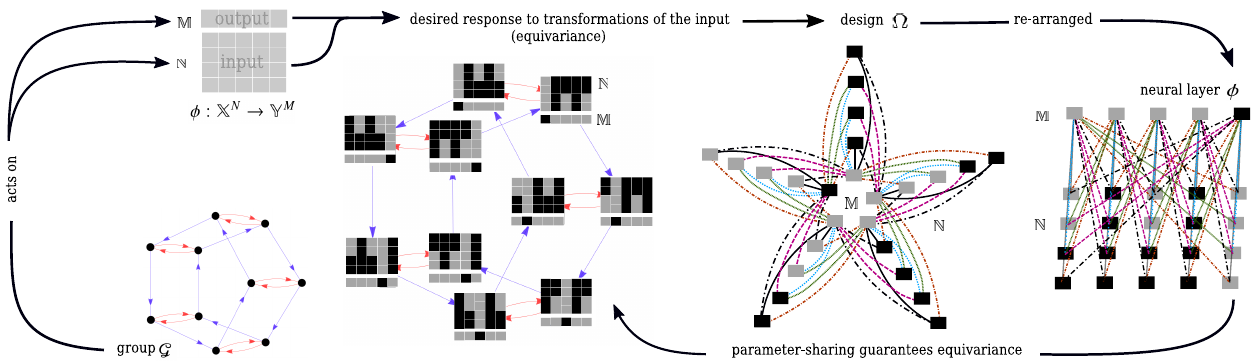}
  \caption{{ Summary: given a group action on input and output of a neural network layer,
      define a parameter-sharing for this layer that is equivariant to these actions.\\
(\textbf{left}) $\GG = \gr{D}_5$ is a Dihedral group, acting on a $4 \times 5$ input image and an output vector of size $5$. $\Nset$ and $\Mset$ denote the index set of input, and output variables respectively.
Here $\GG$ is represented using its Cayley diagram.\\
(\textbf{middle-left})
$\GG$-action for $\gg \in \GG$ is shown for an example input.
$\GG$-action on the input is a combination of circular shifts (blue arrows) 
and vertical flips (red arrows) of the 2D image. $\GG$ acts on the output indices $\Mset$ only through circular shift. A permutation group $\GG\ttt{\Nset,\Mset}$ encodes 
the simultaneous ``action'' of $\GG$ on input and output indices.\\
(\textbf{middle-right}) The structure $\Str$ designed using our procedure, such that its symmetries $\aut(\Str)$ subsumes the permutation group $\GG\ttt{\Nset, \Mset}$.\\
(\textbf{right}) the same structure $\Str$ unfolded to a bipartite form to better show the resulting parameter-sharing
in the neural layer. The layer is equivariant to $\GG$-action:
shifting the input will shift the output of the resulting neural network function,
while flipping the input does not change the output.}
}\label{fig:main}
\end{figure*}

This work is an attempt at answering this question, when our priors are in the form discrete
domain symmetries.
To formalize this type of prior, a family of transformations of input and output to a neural layer 
are expressed as group ``action'' on the input and output.
The resulting neural network is \textit{invariant} to this action, if transformations of the input within that particular family, does not change the output (\eg rotation-invariance).
However, if the output is transformed, in a predictable way, as we transform the input, the neural layer is \textit{equivariant} to the
action of the group. Our goal is to show that parameter-sharing can be used to achieve equivariance to any discrete group action.

Application of group theory in machine learning has been the topic of various works in the past~\citep[\eg][]{kondor2008group,bartok2010models}.
In particular, many probabilistic inference techniques have been extended to graphical models with known symmetry groups~\citep{raedt2016statistical,kersting2009counting,bui2012automorphism,niepert2012markov}.
Deep and hierarchical models have used a variety of techniques to study or obtain representations that isolate
transformations from the 
``content''~\citep[\eg][]{hinton2011transforming,jayaraman2015learning,lenc2015understanding,agrawal2015learning}.
The simplest method of achieving equivariance is through data-augmentation~\citep{krizhevsky2012imagenet,dieleman2015rotation}.  
Going beyond augmentation, several methods directly apply the group-action, in one way or another, 
by transforming the data or \textit{its encodings} using group members~\citep{jaderberg2015spatial,anselmi2013unsupervised,dieleman2016exploiting}.
An alternative path to invariance via harmonic analysis. In particular cascade of wavelet transforms is investigated in~\cite{bruna2013invariant,oyallon2015deep,sifre2013rotation}.
More recently \cite{cohen2016steerable} study steerable filters~\citep[\eg][]{freeman1991design,hel1998common} as a general mean for achieving equivariance in deep networks.
Invariance and equivariance through parameter-sharing is also discussed in several prior works~\citep{cohen2016group,gens2014deep}.

The desirability of using parameter-sharing for this purpose is mainly due to its simplicity and computational efficiency.
However, it also suggests possible directions for \textit{discovering} domain symmetries through regularization schemes. 
Following the previous work on the study of symmetry in deep networks, we rely on group theory and group-actions to formulate
invariances and equivariances of a function.
Due to discrete nature of parameter-sharing, our treatment here is limited to permutation groups.
Action of a permutation group $\GG$ can model discrete transformations of a set of variables, such as translation and $90^{\circ}$ rotation of pixels around any center in an image.
If the output of a function transforms with a $\GG$-action as we transform its input with a different $\GG$-action, the function is equivariant
with respect to action of $\GG$. For example, in a convolution layer, as we translate the input,
 the feature-maps are also translated.
If the output does not transform at all, the function is invariant
to the action of $\GG$. Therefore, invariance is a special equivariance.
In this example, different translations correspond to the action of different members of $\GG$.

The \textit{novelty} of this work is its focus on the ``model symmetry'' as a gateway to
equivariance. This gives us new theoretical guarantees for a ``strict'' notion of equivariance in neural networks. 
The core idea is simple: consider a colored bipartite graph $\Str$ representing a neural network layer.
Edges of the same color represent tied parameters.
This \textit{neural network layer as a function is equivariant
to the actions of a given group $\GG$ (and nothing more) iff the action of $\GG$ is the symmetry group of $\Str$} --
\ie there is a simple bijection between
parameter symmetries and equivariences of the corresponding neural network.

The problem then boils down to \textbf{designing} colored bipartite graphs with 
 given symmetries, which constitutes a major part of this paper.
\cref{fig:main} demonstrates this idea.\footnote{Throughout this paper, since we deal with finite sets, we use circular shift and circular convolution instead of shift and convolution. The two can be made identical with zero-padding of the input.}


For the necessary background on group theory see the Appendix.
In the following, \cref{sec:main} formalizes equivariance wrt discrete group action.
\cref{sec:layer} relates the model symmetries a neural layer to its equivariance.
\cref{sec:design} then builds on this observation to introduce two procedures for parameter-sharing that achieves a desirable equivariance.
Here, we also see how group and graph convolution as well as deep-sets become special instances in our parameter-sharing procedure,
which provides new insight and improved design in the case of group convolution.
Where input and output of the layer have a one-to-one mapping, we see that the design problem reduces a well-known problem in combinatorics.

\section{Group Action and Equivariance}\label{sec:main}
%


Let $\xx = [x_1,\ldots,x_N] \in \XX^{N}$  denote a set of variables and $\GG = \{\gg\}$ be a finite group. The discrete action of $\GG$ on $\xx$ is in the form of permutation of indices in $\Nset = \{1,\ldots,N\}$. 
This group is a subgroup of the \textit{symmetric group} $\gr{S}\ttt{\Nset}$; the group of all $N!$ permutations of $N$ objects. We use $\Nvec = [1,\ldots,N]$ to denote the ordered counterpart to $\Nset$ and the $\GG$-action on this vector $\gg \Nvec \defeq [\gg 1, \ldots, \gg N]$ is a simple permutation. Using $\xx_{\Nvec}$ to denote $\xx$, the discrete action of $\gg \in \GG$ on $\xx \in \XX^{N}$ is given by $\gg \xx_{\Nvec} \defeq \xx_{\gg \Nvec}$.

$\GG$-action on $\Nset$ is a permutation group that is not necessarily isomorphic to $\GG$ itself.
$\GG\ttt{\Nset} \leq \GG$ captures the structure of $\GG$ when it acts on $\Nset$.
We use $\gg\ttt{\Nset}$ to denote the image of $\gg\in\GG$ in $\GG\ttt{\Nset}$. 
$\GG$-action is \textbf{faithful} iff two groups are isomorphic $\GG \cong \GG\ttt{\Nset}$ -- that
is $\GG$-action preserves its structure.
In this case, each $\gg \in \GG$ maps to a \textit{distinct} permutation $\gg \Nvec \neq \gg' \Nvec \forall \gg, \gg' \in \GG$.
Given any $\GG$-action on $\Nset$ we can efficiently obtain $\GG\ttt{\Nset}$; see Appendix.

\begin{mdframed}[style=MyFrame2]
\begin{example}[Cyclic Group]
  Consider the cyclic group $\GG = \mathbb{Z}_6$ and define its action on $\xx \in \Re^3$ by
  defining it on the index set $\Nset = \{1,2,3\}$ as
  $\gg n \defeq \gg + n \mod 3\, \forall \gg \in \mathbb{Z}_6$.
  This action is not faithful. For example, the action of $\gg=\gr{1}$ and $\gg = \gr{4}$ result in the same permutations of variables in $\xx$; \ie single-step of circular shift to the right.
  With the above action, the resulting permutation group $\GG\ttt{\Nset}$ is isomorphic to $\mathbb{Z}_3 < \mathbb{Z}_6$.
  
  Now consider the same group $\GG = \mathbb{Z}_6$ with a different action on $\Nset$: $\gg n \defeq \gg - n \mod 3\, \forall \gg \in \mathbb{Z}_6$, where we replaced $(+)$ with $(-)$.
  Let $\tilde{\GG}\ttt{\Nset}$ be the resulting permutation group. Here again $\tilde{\GG}\ttt{\Nset} \cong \mathbb{Z}_3$.
  Although isomorphic, $\tilde{\GG}\ttt{\Nset} \neq {\GG}\ttt{\Nset}$, as they are different permutation groups of $\Nset$.
\end{example}
\end{mdframed}

Consider the function $\phis: \XX^N \to \YY^M$
and let $\GG\ttt{\Nset}$ and $\GG\ttt{\Mset}$ be the action of $\GG$ on input/output
index sets $\Nset$ and $\Mset$.
\begin{definition}
The joint permutation group $\GG\ttt{\Nset, \Mset}$ is a sub-direct product (or pairing) of $\GG\ttt{\Nset}$ and $\GG\ttt{\Mset}$
$$\GG\ttt{\Nset, \Mset} = \GG\ttt{\Nset} \odot \GG\ttt{\Mset} \defeq \{(\gg\ttt{\Nset}, \gg\ttt{\Mset}) \mid \gg \in \GG\}.$$
\end{definition}

We are now ready to define equivariance and invariance.
$\phis(\cdot)$ is $\GG\ttt{\Nset, \Mset}$-\textbf{equivariant} iff 
\begin{align}
  \label{eq:equivariance}
  \gg\ttt{\Nset} \phis(\xx) = \phis(\gg\ttt{\Mset} \xx) \quad \forall \xx \in \XX^{N}, (\gg\ttt{\Nset}, \gg\ttt{\Mset}) \in \GG\ttt{\Nset, \Mset}
\end{align}
Moreover, if $\GG\ttt{\Mset} = \{\gr{e}\}$ is trivial, we have
$$\gg\ttt{\Nset} \phis(\xx) = \phis(\xx) \quad \forall \xx \in \XX^{N}, \gg\ttt{\Nset} \in \GG\ttt{\Nset}$$
and $\phis(\cdot)$ is $\GG\ttt{\Nset}$-\textbf{invariant}.

$\gg\ttt{\Nset}$ and $\gg\ttt{\Mset}$ can also be represented using \textbf{permutation matrices}
$\mathbf{G}\ttt{\Nset} \in \{0,1\}^{N \times N}$, 
and $\mathbf{G}\ttt{\Mset} \in \{0,1\}^{M \times M}$.
Equivariance relation of \cref{eq:equivariance} then becomes
\begin{align}\label{eq:equivariance_matrix}
\mathbf{G}\ttt{\Mset} \phis(\xx) = \phis(\mathbf{G}\ttt{\Nset} \xx) \, \forall \xx \in \XX^{N}, (\mathbf{G}\ttt{\Nset}, \mathbf{G}\ttt{\Mset}) \in \GG\ttt{\Nset, \Mset}
\end{align}

The following observation shows that the subgroup relationship affects equivariance and invariance.
\begin{observation}\label{obs:1}
  If the function $\phis: \XX^{N} \to \YY^{M}$ is $\GG\ttt{\Nset, \ttt{\Mset}}$-equivariant,
  then it is also $\HH\ttt{\Nset, \ttt{\Mset}}$-equivariant for any permutation group $\HH\ttt{\Nset, {\Mset}} < \GG$. 
\end{observation} 

\begin{mdframed}[style=MyFrame2]
\begin{example}[Reverse Convolution]
  Consider the cyclic group $\GG = \mathbb{Z}_6$ and for $\gg \in \GG$, define the action on $\Nset = \{1,2,3\}$ to be $\gg n \defeq \gg + n \mod 3$.
  Also let its action on $\Mset = \{1,\ldots,6\}$ be $\gg m \defeq \gg - n \mod 6$.
  In other words, $\GG$-action on $\Nset$ performs circular shift to the \textit{right} and its action on $\Mset$ shifts variables to the \textit{left}.
  Examples of the permutation matrix representation for two members of $\GG\ttt{\Nset}$ and $\GG\ttt{\Mset}$
  are 
   \begin{align*}
    \gr{2}\ttt{\Nset} =  \left ( \begin{smallmatrix}
    0 & 1 & 0 \\
    0 & 0 & 1 \\
    1 & 0 & 0 \\
  \end{smallmatrix}\right)
     \qquad
\gr{2}\ttt{\Mset} =   \left (\begin{smallmatrix}
  0 & 0 & 1 & 0 & 0 & 0 \\
  0 & 1 & 0 & 0 & 0 & 0 \\
  1 & 0 & 0 & 0 & 0 & 0 \\  
  0 & 0 & 0 & 0 & 0 & 1 \\
  0 & 0 & 0 & 0 & 1 & 0 \\
  0 & 0 & 0 & 1 & 0 & 0 \\  
  \end{smallmatrix} \right)
   \end{align*}
   corresponding to right and left shift on vectors of different lengths.
     Now consider the function $\phis: \Re^{N} \to \Re^{M}$
     \begin{align*}
       \phis_\WW(\xx) = \WW \xx \quad \WW^{\mathsf{T}} = \left ( \begin{smallmatrix}
  0 & a & b & 0 & a & b \\
  a & b & 0 & a & b & 0 \\
  b & 0 & a & b & 0 & a \\ 
  \end{smallmatrix}\right ) \quad \forall a,b \in \Re
     \end{align*}
     Using permutation matrices one could check the equivariance condition  \cref{eq:equivariance_matrix} for this function. We can show that $\phis$ is equivariant
     to $\GG\ttt{\Nset, \Mset}$. Consider $\gr{2} \in \mathbb{Z}_6$ and its
     images $\gr{2}\ttt{\Nset} \in \GG\ttt{\Nset}$ and
     $\gr{2}\ttt{\Mset} \in \GG\ttt{\Mset}$. L.h.s. of \cref{eq:equivariance_matrix} is
     \begin{align*}
       \gr{2}\ttt{\Mset}\phis_\WW(\xx) = \left (\begin{smallmatrix}
  0 & 0 & 1 & 0 & 0 & 0 \\
  0 & 1 & 0 & 0 & 0 & 0 \\
  1 & 0 & 0 & 0 & 0 & 0 \\  
  0 & 0 & 0 & 0 & 0 & 1 \\
  0 & 0 & 0 & 0 & 1 & 0 \\
  0 & 0 & 0 & 1 & 0 & 0 \\  
\end{smallmatrix} \right) \left (\begin{smallmatrix}
  0 & a & b \\
  a & b & 0 \\
  b & 0 & a \\
  0 & a & b \\
  a & b & 0 \\
  b & 0 & a \\
\end{smallmatrix}\right ) \xx  =  \left (\begin{smallmatrix}
   b & 0 & a\\
   0 & a & b\\
   a & b & 0\\
   b & 0 & a\\
   0 & a & b\\
   a & b & 0\\
 \end{smallmatrix}\right ) \xx
     \end{align*}

     which is equal to its r.h.s.
     \begin{align*}
       \phis_\WW(\gr{2}\ttt{\Nset} \xx) =  \left (\begin{smallmatrix}
  0 & a & b \\
  a & b & 0 \\
  b & 0 & a \\
  0 & a & b \\
  a & b & 0 \\
  b & 0 & a \\
  \end{smallmatrix}\right ) \left(\begin{smallmatrix}
    0 & 1 & 0 \\
    0 & 0 & 1 \\
    1 & 0 & 0 \\
  \end{smallmatrix}\right ) \xx = \left (\begin{smallmatrix}
   b & 0 & a\\
   0 & a & b\\
   a & b & 0\\
   b & 0 & a\\
   0 & a & b\\
   a & b & 0\\
\end{smallmatrix}\right ) \xx
     \end{align*}
     for any $\xx$. One could verify this equality for all $\gg \in \mathbb{Z}_6$.

     Now consider the group $\HH\ttt{\Nset, \Mset} < \GG\ttt{\Nset, \Mset}$,
     where $\HH\ttt{\Nset} = \GG\ttt{\Nset}$ and members of 
      $\HH\ttt{\Mset} = \{\gr{0}, \gr{2}, \gr{4}\}$,
      perform left circular shift of length $0, 2$ and $4$.
      It is easy to see that $\HH\ttt{\Nset, \Mset} \cong \mathbb{Z}_3$.
      Moreover since $\HH\ttt{\Nset, \Mset} < \GG\ttt{\Nset, \Mset}$, $\phis(\cdot)$ above is
      $\HH\ttt{\Nset, \Mset}$-equivariant as well. However, one prefers to characterize the equivariance
      properties of $\phis$ using $\GG\ttt{\Nset, \Mset}$ rather than $\HH\ttt{\Nset, \Mset}$.
\label{example:circ2}
\end{example}
\end{mdframed}

The observation above suggests that $\GG\ttt{\Nset, \Mset}$-equivariance is not restrictive enough.
As an extreme case, a constant function $\phis(\xx) = \mathbf{1}$ is equivariant to any
permutation group $\GG\ttt{\Nset, \Mset} \leq \gr{S}\ttt{\Nset} \times \gr{S}\ttt{\Mset}$.
In this case equivariance of $\phis$ with respect to a particular $\GG\ttt{\Nset, \Mset}$ is not very informative to us. To remedy this, we define a more strict notion of equivariance.
\begin{definition}
we say a function $\phis: \XX^{N} \to \YY^{M}$ is \textbf{uniquely $\GG$-equivariant} iff it is
$\GG$-equivariant and it is ``not'' $\HH$-equivariant for any $\HH > \GG$.
\end{definition}

\section{Symmetry Groups of a Network}\label{sec:layer}
Given a group $\GG$, and its discrete action through $\GG\ttt{\Nset, \Mset}$,
we are interested in defining parameter-sharing
schemes for a parametric class of functions that guarantees their unique $\GG\ttt{\Nset, \Mset}$-equivariance.
We start by looking at a single neural layer and
relate its unique $\GG\ttt{\Nset, \Mset}$-equivariance to the 
symmetries of a colored multi-edged bipartite graph that defines parameter-sharing.
We then show that the idea extends to multiple-layers.

\begin{definition}
A colored multi-edged bipartite graph $\Str = (\Nset,\Mset, \alpha)$ is a triple, 
where $\Nset$ and $\Mset$ are its two sets of nodes, and $\alpha: \Nset \times \Mset \to 2^{\{1,\ldots,C\}}$
is the edge function that assigns multiple edge-colors from the set $\{1,\ldots,C\}$ to each edge. Non-existing edges receive no color.
\end{definition}

We are interested in the {symmetries} of this structure.
The set of permutations $(\pis\ttt{\Nset}, \pis\ttt{\Mset}) \in \gr{S}\ttt{\Nset} \times \gr{S}\ttt{\Mset}$ of nodes (within each part of the bipartite graph) 
that preserve all edge-colors define the \textbf{Automorphism Group} $\aut(\Str) \leq \gr{S}\ttt{\Nset} \times \gr{S}\ttt{\Mset}$ -- that is $\forall (n,m) \in \Nset \times \Mset$
\begin{align}
  \label{eq:aut_bipartite}
&(\pis\ttt{\Nset}, \pis\ttt{\Mset}) \in \aut(\Str) \; \Leftrightarrow \; \alpha(n,m) = \alpha( (\pis\ttt{\Nset} n, \pis\ttt{\Mset} m)) 
\end{align}

Alternatively, to facilitate the notation, we define the same structure (colored multi-edged bipartite graph) 
as \textbf{a set of binary relations} between $\Nset$
and $\Mset$ -- that is
$\Str = (\Nset, \Mset, \{ \Rel_c\}_{1 \leq c \leq C})$ where each relation is associated with one color 
$\Rel_c = \{(n,m) \mid c \in \alpha(n,m) \forall (n,m) \in \Nset \times \Mset\}$. 
This definition of structure, gives an alternative expression for $\aut(\Str)$
\begin{align}
  \label{eq:aut_bipartite2}
&(\pis\ttt{\Nset}, \pis\ttt{\Mset}) \in \aut(\Str) \; \Leftrightarrow  \\
&\bigg( (n,m) \in \Rel_c  \; \Leftrightarrow\; (\pis\ttt{\Nset} n, \pis\ttt{\Mset} m) \in \Rel_c \bigg)  \quad \forall c, n, m \notag
\end{align}

The significance of this structure is in that, 
it defines a parameter-sharing scheme in a neural layer, where the same edge-colors
correspond to the same parameters. Consider the function $\phis\defeq [\phi_1,\ldots,\phi_M]: \Re^{N} \to \Re^{M}$
\begin{align}
  \label{eq:layer}
  \phi_{m}(\xx; \thetas, \Str) \defeq \sigma \bigg(\sum_{n} \sum_{c \in \alpha(n,m)} \theta_c x_{n}\bigg) \quad \forall m
\end{align}
where $\sigma: \Re \to \Re$ is a strictly \textit{monotonic} nonlinearity and
$\thetas = [\theta_1,\ldots\theta_c,\ldots,\theta_C]$ is the parameter-vector for this layer.
 
The following key theorem relates the equivariances of $\phis(\cdot; \thetas, \Str)$ to the symmetries of $\Str$.
\begin{mdframed}[style=MyFrame]
\begin{theorem}\label{th:main}
For any $\ww \in \Re^{C}$ s.t., $w_c \neq w_{c'} \forall c,c'$, the function $\phis(\cdot; \thetas, \Str)$ is uniquely $\aut(\Str)$-equivariant.
\end{theorem}
\end{mdframed}
\begin{corollary}\label{cor:main}
  For any $\HH\ttt{\Nset, \Mset} \leq \aut(\Str)$, the function $\phis(\cdot; \thetas, \Str)$ is $\HH\ttt{\Nset, \Mset}$-equivariant.
\end{corollary}
The implication is that to achieve unique equivariance for a given group-action,
we need to define the parameter-sharing using the structure $\Str$ with symmetry group $\GG\ttt{\Nset, \Mset}$.

\begin{mdframed}[style=MyFrame2]
\begin{example}[Reverse Convolution]
  \label{example:circ3}
  Revisiting Example~\ref{example:circ2} we can show that the condition of Theorem~\ref{th:main} holds.
  In this case $\sigmas(\xx) = \xx$ and the parameter-sharing of the matrix $\WW$ is visualized below,
  where we used two different line styles for $a, b \in \Re$.
  \begin{center}
    \includegraphics[width=\linewidth]{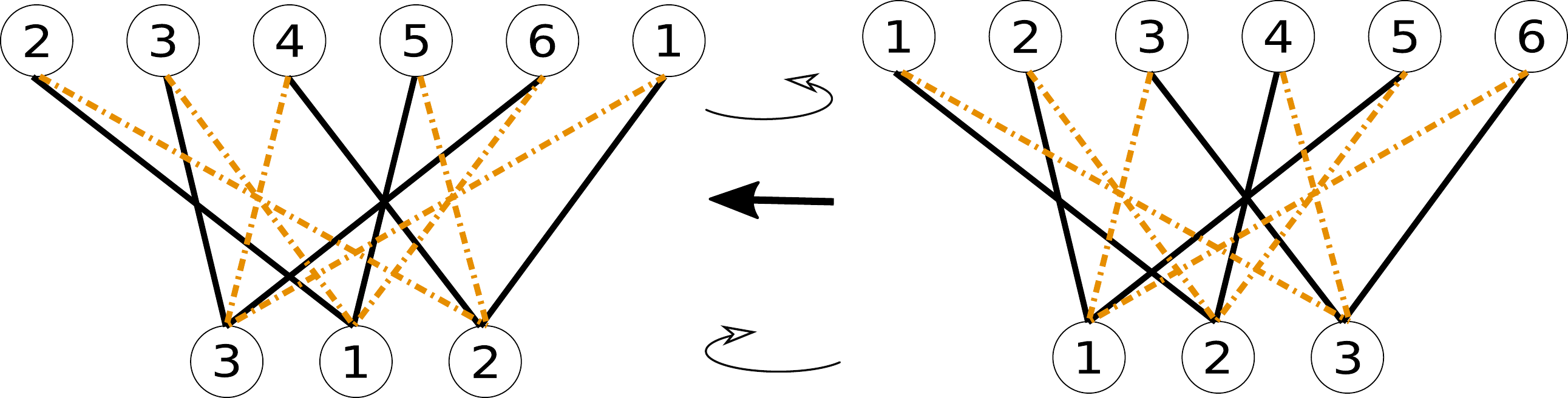}
  \end{center}
In this figure, the
circular shift of variables at the output and input level to the left and right respectively, does not change the
edge-colors. For example in both cases node \circled{1}'s connection to nodes \circled{3}, \circled{6} using dashed-lines is preserved.

Six repetitions of this action produces different permutations corresponding to 
six members of $\GG\ttt{\Nset, \Mset}$. Therefore $\GG\ttt{\Nset, \Mset} \leq \aut(\Omega)$
and according to Corollary~\ref{cor:main}, $\phi(\cdot)$ is  $\GG\ttt{\Nset, \Mset}$ equivariant.
Moreover, using Theorem~\ref{th:design} of the next section, we can prove that these six permutations are the ``only''
edge-color preserving ones for this structure, resulting in \textit{unique equivariance}.
\end{example}
\end{mdframed}

\textbf{Matrix Form.}
To write \cref{eq:layer} in a matrix form, if there are multiple edges between
two nodes $n,m$, we need to merge them.
In general, \textit{by assigning on distinct color to any set in the range of $\alpha: \Nset \times \Mset \to 2^{\{1,\ldots,C\}}$ we can
  w.l.o.g. reduce multiple edges to a single edge.}
In other words we can rewrite $\phis$ using $\WW \in \Re^{M \times N}$
\begin{align}\label{eq:layer_matrix}
\phis(\xx; \thetas; \Str) = \sigmas (\WW \xx)  \quad W_{m,n} = \sum_{c \in \alpha(n,m)} \theta_c  
\end{align}
Using this notation, and due to strict monotonicity of the nonlinearity $\sigma(\cdot)$, Theorem~\ref{th:main} simply states that for all $(\gg\ttt{\Nset},\gg\ttt{\Mset}) \in \aut(\Str)$, $\xx \in \Re^{N}$ and $\WW$ given by \cref{eq:layer_matrix}
\begin{align}\label{eq:commute}
  \mathbf{G}\ttt{\Mset} \WW \xx =  \WW \mathbf{G}\ttt{\Nset} \xx.
\end{align}



\begin{mdframed}[style=MyFrame2]
\begin{example}[Permutation-Equivariant Layer]\label{example:perm_equivariant}
  Consider all permutations of indices $\Nset$ and $\Mset = \Nset$.
  \begin{center}
    \includegraphics[width=.3\linewidth]{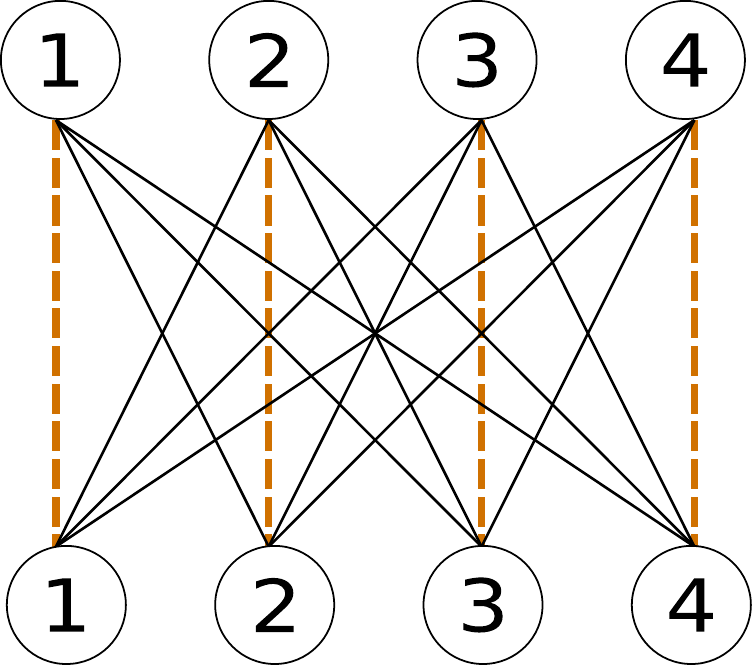}
  \end{center}
  We want to define a neural layer such that all permutations of the input $\gg\ttt{\Nset} \in \GG\ttt{\Nset} = \gr{S}\ttt{\Nset}$ result in the
  same permutation of the output $\gg\ttt{\Mset} = \gg\ttt{\Nset}$.
  Consider the following colored bipartite graph, for a special case where $N = M = 4$.
    It is easy to show that color-preserving permutations of this structure are $\aut(\Str) = \gr{S}\ttt{\Nset} \odot \gr{S}\ttt{\Nset} = \{ (\gg, \gg) \mid \gg \in \gr{S}\ttt{\Nset}\} \cong \gr{S}\ttt{\Nset}$:
On one hand, for $(\pis\ttt{\Nset},\pis\ttt{\Mset}) \in \gr{S}\ttt{\Nset} \times \gr{S}\ttt{\Mset}$, having $\pis\ttt{\Nset} = \pis\ttt{\Mset}$ clearly preserves the colors. On the other hand, if $\pis\ttt{\Nset} \neq \pis\ttt{\Mset}$, there exists $u \in \Nset$ (also in $\Mset$) such that $\pis\ttt{\Nset} u \neq \pis\ttt{\Mset} u$.
Therefore $(\pis\ttt{\Nset}, \pis\ttt{\Mset})$ does not preserve the relation $\Rel = \{(n,n) \mid n \in \Nset\}$ corresponding to dashed edges, and therefore $(\pis\ttt{\Nset}, \pis\ttt{\Mset}) \notin \aut(\Str)$. This proves $\aut(\Str) = \gr{S}\ttt{\Nset} \odot \gr{S}\ttt{\Nset}$.
The function \cref{eq:layer} for this $\Str$ is $$\phis(\xx; \thetas = [\theta_1, \theta_2], \Str) = \sigmas (\theta_1 \mathbf{I} \xx + \theta_2 \mathbf{1} \mathbf{1}^{\mathsf{T}} \xx).$$
\citet{ravanbakhsh2016deep,deep_sets} derive the same permutation equivariant layer,
    by proving the commutativity in \cref{eq:commute}, while here it follows from Corollary~\ref{cor:main}. 
\end{example}
\end{mdframed}

\textbf{Multiple Layers.} For deep networks,
the equivariance of the composition $\phis_{2} \circ \phis_1$ to $\GG$-action
follows from that of individual layer $\phis_{1}:\XX^{N} \to \YY^{M}$ and $\phis_{2}:\YY^{M} \to \mathbb{Z}^{O}$.
Assuming $\phis_1$ is $\GG\ttt{\Nset, \Mset}$-equivariant and $\phis_2$ is $\GG\ttt{\Mset,\mathbb{O}}$-equivariant, where $\GG$-action on $\Mset$ is shared between the two layers, it follows that $\phis_{2} \circ \phis_1$ is $\GG\ttt{\Nset, \mathbb{O}}$-equivariant, where
$\GG\ttt{\Nset, \mathbb{O}} = \GG\ttt{\Nset}\odot \GG\ttt{\mathbb{O}}$.
This is because $\forall \gg \in \GG$ and $\xx \in \XX^{N}$
\begin{align}
  \label{eq:multi_layer}
  \phis_{2}(\phis_{1}(\gg\ttt{\Nset} \xx)) = \phis_{2}(\gg\ttt{\Mset} \phis_{1}(\xx)) = \gg\ttt{\mathbb{O}} \phis_{2}(\phis_{1}(\xx)).
\end{align}



\section{Structure Design}\label{sec:design}
Consider the definition of neural layer \cref{eq:layer} that employs parameter-sharing according to $\Str$.
Given $\GG$-action on $\Nset$ and $\Mset$, we are interested in designing
structures $\Str$ such that $\aut(\Str) = \GG\ttt{\Nset,\Mset}$. According to the Theorem~\ref{th:main},
it then follows that $\phis$ is uniquely $\GG\ttt{\Nset,\Mset}$-equivariant.
Here, we give the sufficient conditions and the design recipe to achieve this. 

For this we briefly review some group properties that are used
in later developments.

\begin{description}
\item[transitivity] We say that $\GG$-action on $\Nset$ is transitive iff $\forall n_1,n_2 \in \Nset$, there exists at least one action $\gg \in \GG$ (or $\gg\ttt{\Nset} \in \GG\ttt{\Nset}$) such that $\gg n_1 = n_2$. 
\item[regularity] The group action is free or \textbf{semi-regular} iff $\forall n_1,n_2 \in \Nset$, there is at most one $\gg \in \GG$ such at $\gg n_1 = n_2$, and the action is \textit{regular} iff it is both transitive and free -- \ie for any pair $n_1,n_2 \in \Nset$, there is uniquely
  one $\gg \in \GG$ such that $\gg n_1 = n_2$. Any free action is also faithful.
  We use a similar terminology for $\GG\ttt{\Nset}$. That is we call $\GG\ttt{\Nset}$ semi-regular
iff $\forall n_1,n_2 \in \Nset$ at most one $\gg\ttt{\Nset} \in \GG\ttt{\Nset}$ moves $n_1$ to $n_2$ and $\GG\ttt{\Nset}$ is regular if this number is exactly one.
\item[orbit] The {orbit} of  $n \in \Nset$ is all the members to which it can be moved, 
$\GG n = \{\gg n \mid \gg \in \GG\}$. The orbits of 
$n \in \Nset$ form an equivalence relation\footnote{$n \sim n' \Leftrightarrow \exists \gg \;\text{s.t.,}\; n = \gg n' \Leftrightarrow n \in \GG n' \Leftrightarrow n' \in \GG n$.}
This equivalence relation \textit{partitions} $\Nset$ into orbits $\Nset = \bigcup_{1 \leq p \leq P}  \GG n_p$, where $n_p$ is an arbitrary \textbf{representative} of the partition 
$\GG n_p \subseteq \Nset$.
Note that the $\GG$-action on $\Nset$ is always transitive on its orbits -- that is for any $n,n' \in \GG n_p$, there is at least one $\gg \in \GG$ such that $n = \gg n'$. Therefore, for a semi-regular $\GG$-action, the action of $\GG$ on the orbits $ \GG n_p \forall 1 \leq p \leq P$ is regular.
\end{description}

\begin{mdframed}[style=MyFrame2]
\begin{example}[Mirror Symmetry]
  Consider $\GG = \mathbb{Z}_2 = \{\gr{e}=\gr{0},\gr{1}\}$ ($1 + 1 = 0$) acting on $\Nset$, where
   the only non-trivial action is defined as flipping the input: $\gr{1}\ttt{\Nset} [1,\ldots,N] = [N,N-1,\ldots,1]$. 
 
   $\GG$ is faithful in its action on $\Nset$, however $\GG\ttt{\Nset}$ is not transitive -- \eg $N$ cannot be moved to $N-1$.
   If $N$ is even, then $\GG$-action is semi-regular. This is because otherwise the element
   in the middle $n = \lceil\frac{N}{2}\rceil$ is moved to itself by two different actions $\gr{e}, \gr{1} \in \GG$. Furthermore, if $N$ is even, $\GG$-action has $\frac{N}{2}$ orbits and $\GG_2$ acts on these orbits regularly. If $N$ is odd, $\GG$-action has $\lceil\frac{N}{2} \rceil$ orbits. However, its action on the orbit of the middle element $\GG\; \lceil \frac{N}{2}\rceil$ is not regular.
\end{example}
\end{mdframed}

In the following, \cref{sec:dense} proposes a procedure for parameter-sharing in a fully connected layer.
Although simple, this design is dense and does not guarantee ``unique'' of equivariance. 
\cref{sec:sparse} proposes an alternative design with sparse connections that in some settings
ensures unique equivariance. 
\cref{sec:multiple_channels} investigates the effect of having multiple input and output channels in the neural layer and \cref{sec:2closed} studies a special case of $\GG\ttt{\Nset} = \GG\ttt{\Mset}$, where input and output indices
have a one-to-one mapping.

\subsection{Dense Design}\label{sec:dense}
Consider a complete bipartite graph with $\Nset$ and $\Mset$ as its two parts and edges $(n,m) \in \Nset\times \Mset$.
The action of $\GG\ttt{\Nset,\Mset}$ partitions the edges into orbits $\{\GG\ttt{\Nset,\Mset} (n_p,m_q)\}_{n_p,m_q}$, where $(n_p,m_q)$ is a representative
edge from an orbit. Painting each orbit with a different color gives 
\begin{align}\label{eq:dense_design}
  \Str = (\Nset, \Mset, \{\Rel_{p,q} = \GG\ttt{\Nset,\Mset} (n_p,m_q)\}).
\end{align}
Therefore two edges $(n,m)$ and $(n',m')$ have the same color iff an action in $\GG\ttt{\Nset,\Mset}$ moves one edge to the other.

\begin{mdframed}[style=MyFrame]
\begin{proposition}\label{th:dense_design}
$\GG\ttt{\Nset,\Mset} \leq \Str$ for $\Str$ of \cref{eq:dense_design}.
\end{proposition}
\end{mdframed}
\begin{corollary}\label{cor:dense_design}
  $\phis(\cdot; \thetas, \Str)$, for structure \cref{eq:dense_design}, is equivariant to $\GG\ttt{\Nset,\Mset}$.
\end{corollary}

\begin{mdframed}[style=MyFrame2]
\begin{example}[Nested Subsets and Wreath Product]
  The permutation-equivariant layer that we saw in Example~\ref{example:perm_equivariant}
  is useful for defining neural layers for set structure. If our data-structure is
  in the form of nested subsets, then we require equivariance to permutation of variables
  within each set as well as permutation of subsets. Here, we show how to use our dense
  design for this purpose.

  We use a special indexing for the input to better identify the exchangeability of variables.
  We assume $D$ subsets, each of which has $d$ variables
  $\xx = [x_{1,1},\ldots, x_{1,d},x_{2,1},\ldots, x_{D,d}]$.

  The group of our interest is
  the \textbf{wreath product} $\gr{S}_d \wr \gr{S}_D$. \textit{This type of group product can be used to build
    hierarchical and nested structures with different type of symmetries at each level.}
  Nesting subsets corresponds to the most basic form of such hierarchical constructions.
  We use $(n,n')$ to index input variables and $(m,m')$ for output variables.

  The following figure shows the resulting parameter-sharing for an example with $D=2$, $d=3$.
  \begin{center}
    \includegraphics[width=.6\linewidth]{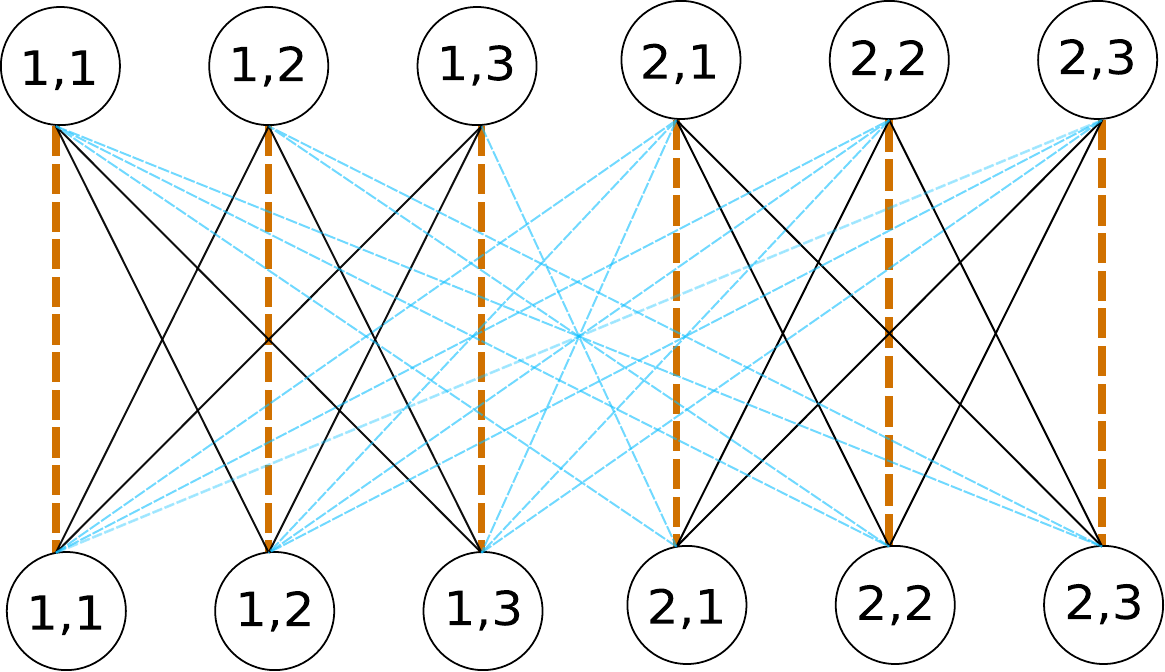}
  \end{center}
    How did we arrive at this structure $\Str$?
    Recall Our objective is to define parameter-sharing so that $\phi_{\WW}: \Re^{dD} \to \Re^{dD}$ is equivariant to the action of $\GG = \gr{S}_d \wr \gr{S}_D$ -- \ie permutations within sets at two levels.
  This group-action identifies three partitions of edges (seen in the figure):
  I) $((n,n'), (n,n')) \forall n,n'$ connects each variable to its counterpart (dashed orange);
  II) $((n,n'), (n,m')) \forall n, n'\neq m'$ connects each variable to other variables within the same subset;
  III) $((n,n'), (m,m')) \forall n \neq m$ is the set of edges from one subset to another.
  According to the Corollary~\ref{cor:dense_design} this parameter-sharing guarantees equivariance.

\end{example}
\end{mdframed}

This fully-connected design is useful when the group $\GG\ttt{\Nset,\Mset}$ is large; for example
when dealing with $\gr{S}_{\Nset}$. However, for smaller groups it could be very 
 inefficient in practice,
 as sometimes we can achieve equivariance through a sparse structure $\Omega$.
As an example, consider the 2D circular convolution layer. 
It is easy to show that according to this design, the convolution filter will be the same size as the input image.
While this achieves the desirable equivariance, it is inefficient and does not generalize as well as a convolution layer with small filters.
Moreover, the dense design does not guarantee ``unique'' equivariance. We next show under some
conditions on $\GG\ttt{\Nset, \Mset}$ the sparse design can produce this stronger guarantee.

\subsection{Sparse Design}\label{sec:sparse}
Our sparse construction uses orbits and symmetric generating sets:
\begin{itemize}
\item Let us denote the orbits of $\GG$-action on $\Mset$ and $\Nset$ by $\{\GG n_p \mid 1 \leq p \leq P\}$
and $\{ \GG m_q \mid 1 \leq q \leq Q\}$ respectively, where $P$ and $Q$ are the total number of orbits and
$n_p$,$m_q$ are (arbitrary) representative members of orbits $\GG n_p$, $\GG m_q$ respectively.
Note that in contrast to previous section, here we are considering the orbit of variables rather than the edges.
\item The set $\gr{A} \subseteq \GG$ is called the generating set of $\GG$ ($<\gr{A}> = \GG$), 
iff every member of $\GG$ can be expressed as a combination of members of $\gr{A}$. 
If the generating set is closed under inverse $\aa \in \gr{A} \Rightarrow \aa^{-1} \in \gr{A}$ we call it a \textbf{symmetric generating set}.
\end{itemize}

Define the structure $\Str$ as 
\begin{align}
  \label{eq:str_general}
  &\Str = (\Nset, \Mset, \{\Rel_{p,q,\aa}\}_{1 \leq p \leq P,\; 1 \leq q \leq Q, \aa \in \gr{A}}) \notag \\
&\Rel_{p,q,\aa} = \{( \gg\ttt{\Nset} \aa n_p, \gg\ttt{\Nset} m_q) \mid (\gg\ttt{\Nset},\gg\ttt{\Mset}) \in \GG\ttt{\Nset,\Mset}\}.
\end{align}

In words, we have one color per each combination of orbits ($p$, $q$) and members
of the generating set $\aa \in \gr{A}$. 
The following theorem relates the symmetry group of this structure to $\GG$.

\begin{mdframed}[style=MyFrame]
\begin{theorem}\label{th:design}
  $\GG\ttt{\Nset,\Mset} \leq \aut(\Str)$ for $\Str$ of \cref{eq:str_general}.
  Moreover if $\GG\ttt{\Nset}$ and $\GG\ttt{\Mset}$ are both semi-regular, then $\GG\ttt{\Nset,\Mset} = \aut(\Str)$.
\end{theorem}
\end{mdframed}
Note that this result holds for any choice of a symmetric generating set $\gr{A}$ in defining $\Str$.
Therefore, in designing sparse layers, one seeks a minimal $\gr{A}$.
\begin{corollary}\label{cor:str_general}
The function $\phis(\cdot, \ww, \Str)$, using the structure \cref{eq:str_general} is $\GG\ttt{\Nset, \Mset}$-equivariant. 
If $\GG\ttt{\Nset}$ and $\GG\ttt{\Mset}$ are semi-regular, this function is ``uniquely'' $\GG\ttt{\Nset, \Mset}$-equivariant.  
\end{corollary}

Now, assuming $\GG$-action is semi-regular on both $\Nset$
and $\Mset$, using (arbitrarily chosen) representatives $\{n_p\}_{1\leq p \leq P}$ and $\{m_q\}_{1 \leq q \leq Q}$ for 
orbits in $\Nset$ and $\Mset$, we can rewrite the expression \cref{eq:layer} of the structured neural layer for the structure above. Here, components of $\phis = [\phi_1,\ldots,\phi_M]$ are enumerated for $1 \leq q \leq Q, \gg\ttt{\Mset} \in \GG\ttt{\Mset}$:
\begin{align}
  \label{eq:layer_str}
  \phi_{\gg\ttt{\Mset} m_q}(\xx; \thetas) = \sigma \bigg (\sum_{1 \leq p \leq P} \sum_{\aa \in \gr{A}} \theta_{q, p, \aa} x_{\gg\ttt{\Nset} \aa n_p} \bigg )
\end{align}
where $\thetas \in \Re^{P \times Q \times |\gr{A}|}$ is the set of unique parameters,
  and each element $\phi_{\gg\ttt{\Mset} m_{q}}$ depends on
 subset of parameters $\{\theta_{q, p, \aa}\}_{p,\aa}$ identified by $q$ and a subset of inputs $\{x_{\aa, \gg\ttt{\Nset} n_p} \}_{p,\aa}$ identified by $\gg\ttt{\Nset}$. 



 \begin{mdframed}[style=MyFrame2]
\begin{example}[Dihedral Group of \cref{fig:main}]
In the example of \cref{fig:main}, the number of orbits 
of $\GG$-action on $\Nset$ is $P=2$ and for $\Mset$ this is $Q=1$.
The symmetric generating set is the generating set that is used in the
Cayley diagram, with the addition of inverse shift (inverse of the blue arrow).
We then used \cref{eq:str_general} to build the structure of \cref{fig:main} (right).
\end{example}

\begin{example}[Reverse Convolution]
  The parameter-sharing structure of reverse convolution in Examples~\ref{example:circ2}
  and ~\ref{example:circ3} is produced using our sparse design.
  In these examples, both $\GG\ttt{\Nset}$ and $\GG\ttt{\Mset}$ are
  regular. Therefore the proposed parameter-sharing provides \textit{unique} equivariance.
\end{example}
\end{mdframed}

\subsection{Multiple Channels}\label{sec:multiple_channels}
In this section, we extend our results to  multiple input and output channels.
Up to this point, we considered a neural network layer $\phis: \Re^{N} \to \Re^{M}$.
Here, we want to see how to achieve $\GG\ttt{\Nset, \Mset}$-equivariance for
$\phis: \Re^{N \times K} \to \Re^{M \times K'}$,
where $K$ and $K'$ are the number of input and output channels.

First, we extend the action of $\GG$ on $\Nset$ and $\Mset$ 
to $\Nset^{K} = [\underbrace{\Nset, \ldots, \Nset}_{K \text{times}}]$ as well as $\Mset^{K'}$,
to accommodate multiple channels.
For this, simply repeat the $\GG$-action on each component.  
$\GG$-action on multiple input channels is equivalent
to sub-direct product $\underbrace{\GG\ttt{\Nset} \odot \ldots \odot \GG\ttt{\Nset}}_{K \text{times}} \cong \GG\ttt{\Nset}$. The same applies to $\GG\ttt{\Mset}$.
  
This repetition, multiplies the orbits of $\GG\ttt{\Nset}$, one for each channel, so that instead of having $P$
and $Q$ orbits on the input $\Nset$ and output $\Mset$ sets, we have $K \times P$ and $K' \times Q$ orbits on the input
$\Nset^K$ and output $\Mset^{K'}$.
This increases the number of parameters by a factor of $K \times K'$.

The important implication is that, \textbf{orbits and multiple channels are treated identically by both dense and sparse designs}.

\begin{mdframed}[style=MyFrame2]
\begin{example}[Group Convolution]
  The idea of group-convolution is studied by \citet{cohen2016group}; see also~\citep{gconv}.
  The following claim relates the function of this type of layer to our sparse design.
  \begin{claim}\label{claim:group_conv}
    Under the following conditions the neural layer \cref{eq:layer} using our sparse design  \cref{eq:str_general} performs group convolution:
    I) there is a bijection between the output and $\GG$ (\ie $\Mset = \GG$) and; II) $\GG\ttt{\Nset}$ is transitive.
  \end{claim}
  This also identifies the limitations of group-convolution even in the setting where $\Mset = \GG$:
  When $\GG\ttt{\Nset}$ is semi-regular and not transitive ($P > 1$), group convolution is not guaranteed to be uniquely equivariant while the sparse parameter-sharing of \cref{eq:str_general}
  provides this guarantee.

  For demonstration consider the following example in \textbf{equivariance to mirror symmetry}.
  \begin{center}
  \includegraphics[width=.7\linewidth]{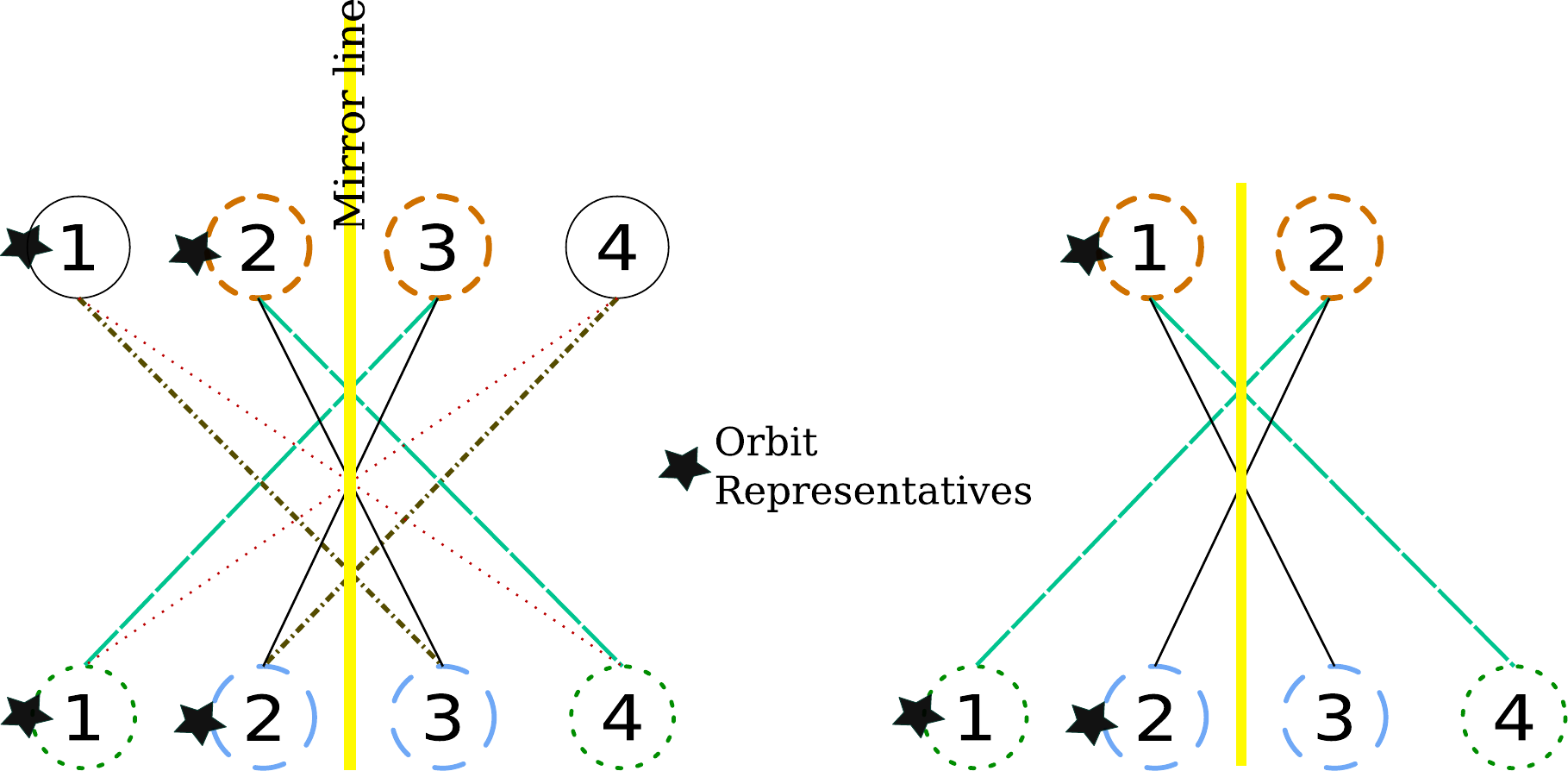}  
  \end{center}
  This figure shows the bipartite structure for $\GG = \mathbb{Z}_2 = \{\gr{0},\gr{1}\}$ and $\gr{A} = \{\gr{1}\}$. $\GG$-action is horizontal flip of the input and the output.
  On the \textit{right}, $\Mset = \GG$ while on the \textit{left}
  $\GG\ttt{\Mset}$-action has two orbits. Orbits are identified by line-style and color of the circles.
  In a neural layer with this parameter-sharing, when we flip the input variables (around the mirror line) the output is also flipped.
  
  The representatives in each orbit on $\Nset$ and $\Mset$ is identified with a star. Note that each combination of orbits $p$ and $q$ has a parameter of its own, identified with different edge-styles. While this construction guarantees ``unique'' $\GG$-equivariance, if instead we use the same parameters across orbits (as suggested by the original group convolution) we get the parameter-sharing of the figure below \textit{middle}.
  \begin{center}
  \includegraphics[width=.9\linewidth]{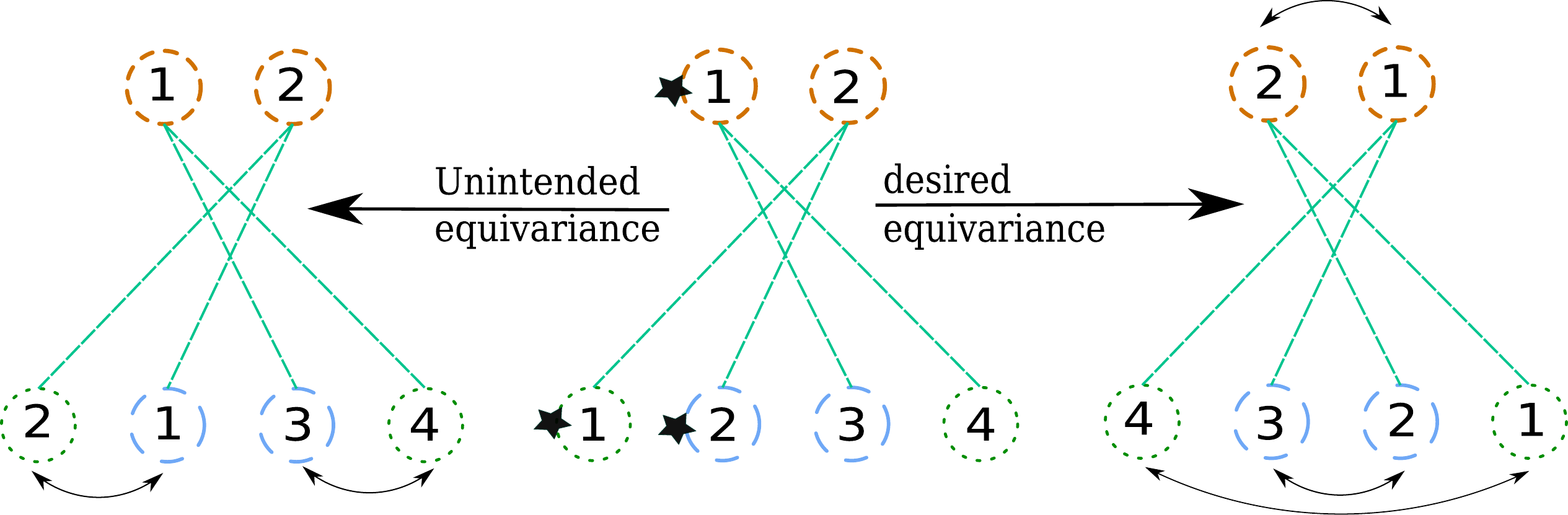}  
\end{center}
In this case, the resulting neural layer has the desired equivariance (\textit{right}).
However, it is equivariant to the action of a larger group $\GG\ttt{\Nset, \Mset} \cong \mathbb{Z}_2 \times \mathbb{Z}_2 > \mathbb{Z}_2$, in which $\gr{1}$ in the second $\mathbb{Z}_2$ group exchanges variables across the orbits on $\Nset$ (\textit{left} in figure above).
\end{example}
\end{mdframed}

\subsection{$\GG\ttt{\Nset} = \GG\ttt{\Mset}$}\label{sec:2closed}\label{sec:semi}
In semi-supervised and un-supervised applications, we often need to produce a single output $y_n$  for each input $x_n \forall n \in \Nset$ -- that is $\Nset = \Mset$.
We can ensure this by having a relation $\Rel_{c^*} = \{(n,n) \mid n \in \Nset\}$ in $\Str$ that guarantees any $(\pis\ttt{\Nset}, \pis\ttt{\Mset}) \in \aut(\Str)$ applies the same permutation to $\Nset$ and 
$\Mset = \Nset$ -- \ie $\pis\ttt{\Nset} = \pis\ttt{\Mset}$. The resulting structure $\Str=(\Nset,\Nset, \{\Rel_c\}_{1 \leq c \leq C} \cup  \{\Rel_{c^*}\}\}$ can be also interpreted as a 
colored multi-edged \textit{directed} graph (digraph). This is because we can collapse the two parts by identifying $n \in \Nset$ with $n \in \Mset$.

Therefore, the symmetry-group of the original bipartite structure, is isomorphic to symmetry group of a colored multi-edged \textit{di}graph on $\Nset$. 
Achieving unique $\GG$-equivariance then reduces to answering the following question:
\textit{when could we express a permutation group $\GG \leq \gr{S}\ttt{\Nset}$ as the symmetry group $\aut(\Str)$ of a colored {multi-edged} digraph with $N$ nodes?}

This problem is well-studied under the class of \textbf{concrete representation problems}~\citep{babai1994automorphism}.
Permutation groups $\GG$ that can be expressed in this way are called \textbf{2-closed groups}~\citep{wielandt1969permutation}. 
The recipe for achieving $\GG\ttt{\Nset} \leq \aut(\Str)$
is similar to our dense construction of \cref{sec:dense}\footnote{In a fully connected digraph,
  the edges that belong to the same orbit by $\GG$-action on $\Nset \times \Nset$, receive the same color.}
The 2-closure $\bar{\GG\ttt{\Nset}}$ of a group $\GG\ttt{\Nset}$ is then, the greatest permutation group $\bar{\GG\ttt{\Nset}} \leq \gr{S}\ttt{\Nset}$ with the same orbit on $\Nset \times \Nset$ as $\GG\ttt{\Nset}$.
It is known that for example semi-regular permutation groups are 2-closed $\bar{\GG\ttt{\Nset}} = \GG\ttt{\Nset}$.
This result also follows a corollary of our Theorem~\ref{th:design} for sparse design of \cref{eq:str_general}. 

\begin{mdframed}[style=MyFrame2]
\begin{example}[Equivariance to $\times 90^\circ$ Rotations]\label{fig:neqm}
  Figure below compares the digraph representation of $\Str$ produced using (\textit{left}) our sparse design, and (\textit{right}) our dense design.
  \begin{center}
  \includegraphics[width=.9\linewidth]{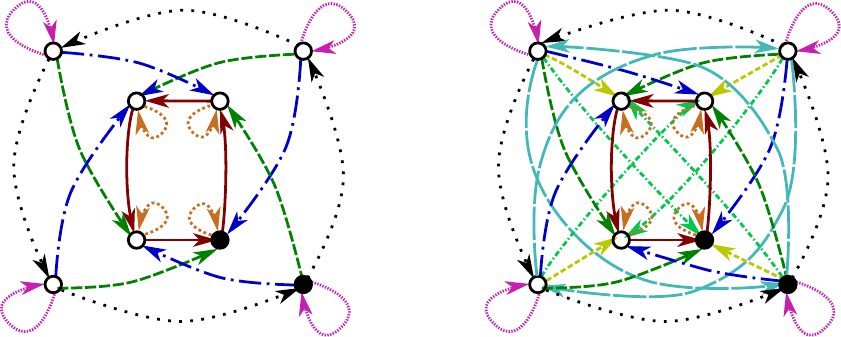}  
  \end{center}
  
Multiples of $\pm 90^\circ$ rotation is produced as the action of cyclic group $\mathbb{Z}_4$ on eight input output variables -- that is $\Nset = \Mset = \{1,\ldots,8\}$.
$\mathbb{Z}_4$-action is semi-regular with two orbits; these orbits the two inner and outer set of four nodes. The representatives of 
each orbit in our sparse design is indicated using filled circles. The generating set consists of $\gr{A} = \{1,3\}$, rotation by $90^\circ$ and its inverse, rotation by $270^\circ$.
Each edge in each of these figures, has a corresponding edge in the opposite direction, within a different relation. To avoid over-crowding the figure,
      we have dropped this edge from the drawing above, unless both edges belong to the same relation.
\end{example}
\end{mdframed}

\begin{mdframed}[style=MyFrame2]
\begin{example}[Graph Convolution]\label{sec:graph_conv}
  Consider the setting where we use the (normalized) adjacency matrix $\mathbf{B}\in \{0,1\}^{N\times N}$ (or Laplacian) of a graph $\Lambda$, to identify parameter-sharing in a neural network layer.
  For a single input/output channel, this is often in the form of
  $\mathbf{A}\xx$, where $\xx \in \Re^{N}$ and $\mathbf{A} = \theta_1 \mathbf{B} + \theta_2 \mathbf{I}$ has different parameters for diagonal and off-diagonal values~\citep[\eg][]{kipf2016semi,bruna2013spectral,henaff2015deep}; for multiple channels see \cref{sec:multiple_channels}.
The following corollary of Theorem~\ref{th:main} identifies the equivariance of $\mathbf{A} \xx$.
\begin{corollary}\label{cor:graph_conv}
  Given the digraph $\Lambda$ and its binary adjacency matrix $\mathbf{B}\in \{0,1\}^{N\times N}$, 
  then $(\theta_1 \mathbf{B} + \theta_2 \mathbf{I}) \xx$ is 
  uniquely equivariant to the symmetry-group of $\Lambda$.
\end{corollary}
Since two graphs on $N$ nodes can have identical symmetries, one implication of this corollary is
that graph-convolution has identical equivariances for graphs with the same symmetry groups.
\end{example}
\end{mdframed}

\section{Conclusion}\label{sec:discussion}
This work is a step towards designing neural network layers with a given
equivariance and invariance properties.
Our approach was to relate the equivariance properties of the neural layer to the symmetries
of the parameter-matrix.

We then proposed two parameter-sharing scheme that achieves equivariance wrt 
any discrete group-action. Moreover under some conditions, we guarantee sensitivity
wrt other group actions. This is important because even a trivial constant function
is invariant to all transformations. It is therefore essential to be able to
draw the line between equivariance/invariance and sensitivity in a function.
To our knowledge, our work presents the first 
results of its kind on guarantees regarding both variance and equivariance with respect to group actions.


\section*{Acknowledgment}
This research is supported in part by DOE grant DESC0011114 and NSF
grant IIS1563887.

\bibliography{iclr2017_conference}
\bibliographystyle{icml2017}
{
\clearpage
\appendix

\section{Proofs}

\begin{proof} of Observation~\ref{obs:1}
$$\gg\ttt{\Mset} \phis(\xx) = \phis(\gg\ttt{\Nset} \xx) \forall \gg \in \GG \Rightarrow \gg\ttt{\Mset} \phis(\xx) = \phis(\gg\ttt{\Mset} \xx) \forall \gg \in \HH \subset \GG.$$
\end{proof}

\begin{proof} of Theorem~\ref{th:main}\\
For unique $\aut(\Str)$-equivariance we need proofs in two directions.
First we show that
\begin{align}\label{proof:960}
(\pis\ttt{\Nset}, \pis\ttt{\Mset}) \in \aut(\Str) \Rightarrow \phis(\xx; \thetas, \Str) = \pis\ttt{\Mset}^{-1}\phis(\pis\ttt{\Nset} \xx; \thetas, \Str)  
\end{align}
which in turn shows that
$\pis\ttt{\Mset} \phis(\xx; \thetas, \Str) = \phis(\pis\ttt{\Nset} \xx; \thetas, \Str)$. Starting from $\pis\ttt{\Mset}^{-1}\phis(\pis\ttt{\Nset} \xx; \thetas, \Str)$ on the r.h.s. of \cref{proof:960} and considering an index $m$ in $\phis = [\phi_1, \ldots, \phi_M]$ we have
\begin{align}
  \phi_{\pis\ttt{\Mset}^{-1} m}(\pis\ttt{\Nset} \xx; \thetas, \Str) &= \sigma \bigg(\sum_{n \in \Nset, c \in \alpha(n, \pis\ttt{\Mset}^{-1} m)} \theta_c x_{\pis\ttt{\Nset} n}\bigg) \notag \\
  &= \sigma \bigg (\sum_{n \in \pis\ttt{\Nset} \Nset, c \in \alpha(\pis\ttt{\Nset}^{-1} n, \pis\ttt{\Nset}^{-1} m)} \theta_c x_{n} \bigg) \notag \\
  &=  \sigma \bigg(\sum_{n \in \Nset, c \in \alpha(n, m)} \theta_c x_{n} \bigg) = \phi_{m}(\xx; \thetas, \Str) \label{proof:981}
 \end{align}
where in arriving at \cref{proof:981} we used the fact that $(\pis\ttt{\Nset},\pis\ttt{\Mset}) \in \aut(\Str) \; \Rightarrow \; \alpha(n,m) = \alpha( (\pis^{-1}\ttt{\Nset} n, \pis^{-1}\ttt{\Mset} m))$.

In the opposite direction we need to show that $\phis(\xx; \thetas, \Str) = \pis\ttt{\Mset}\phis(\pis^{-1}\ttt{\Nset} \xx; \thetas, \Str)\; \forall \xx \in \Re^{N}, \thetas \in \Re^{C}$ only if $(\pis\ttt{\Nset}, \pis\ttt{\Mset}) \in \aut(\Str)$.
\begin{align}
&\phis(\xx; \thetas, \Str) = \pis\ttt{\Mset}\phis(\pis^{-1}\ttt{\Nset} \xx; \thetas, \Str) \; \forall \xx \in \Re^{N}, \thetas \in \Re^{C}  \Rightarrow\\
&\phi_m(\xx; \thetas, \Str) = \phi_{\pis\ttt{\Mset} m}(\pis^{-1}\ttt{\Nset} \xx; \thetas, \Str) \; \forall m, \xx \in \Re^{N}, \thetas \in \Re^{C} \Rightarrow\\
&\forall m, \xx \in \Re^{N}, \thetas \in \Re^{C} \\
&\sum_{n \in \Nset, c \in \alpha(n, m)} \theta_c x_{n} = \sum_{n \in \Nset, c \in \alpha( n, \pis\ttt{\Mset} m)} \theta_c  x_{\pis^{-1}\ttt{\Nset} n}  \Rightarrow \label{proof:984} \\
&\sum_{n \in \Nset, c \in \alpha(n, m)} \theta_c x_{n} = \sum_{n \in \Nset, c \in \alpha(\pis\ttt{\Nset} n, \pis\ttt{\Mset} m)} \theta_c  x_{n} \label{proof:985} 
\end{align}
where \cref{proof:984} follows from monotonicity of $\sigma: \Re \to \Re$.
We need to show that this final equality $\forall m, \xx \in \Re^{N}, \thetas \in \Re^{C}$
implies that $\alpha(\pis\ttt{\Nset} n, \pis\ttt{\Mset} m) = \alpha(n, m)$, which in turn, according to \cref{eq:aut_bipartite} means $(\pis\ttt{\Nset},\pis\ttt{\Mset})  \in \aut(\Str)$.

We prove $\alpha(\pis\pis\ttt{\Nset} n, \pis\pis\ttt{\Mset} m) = \alpha(n, m)$ by contradiction: assume $\alpha(\pis\pis\ttt{\Nset} n^*, \pis\pis\ttt{\Mset} m^*) \neq \alpha(n^*, m^*)$ for some $n^*,m^*$. 

Since $\alpha(\pis\ttt{\Nset} n^*, \pis\ttt{\Mset} m^*) \neq \alpha(n^*, m^*)$, we can
w.l.o.g. assume $\exists c^* \in \alpha(n^*,m^*)\; \text{s.t.}\; c^* \notin \alpha(\pis n^*, \pis m^*)$ (the reverse direction, where $c^* \in \alpha(\pis\ttt{\Nset} n^*, \pis\ttt{\Mset} m^*) \wedge c^* \notin \alpha(n^*,  m^*)$ is similar).
We show that an assignment of $\xx\in \Re^{N}$ and $\theta \in \Re^{C}$ contradicts \cref{proof:985}.
For this, define $\xx$ such that $x_{n} = \delta(n, n^*)$, is non-zero only at index $n^*$. Moreover, assigning $\theta_c = \delta(c, c^*)$ the r.h.s. of \cref{proof:985} is
$\sum_{n \in \Nset, c \in \alpha(\pis\ttt{\Nset} n, \pis\ttt{\Mset} m^*)} \theta_c  x_{n} = 0$
while the l.h.s. is
$\sum_{n \in \Nset, c \in \alpha(m, n)} \theta_c x_{n} = \theta_{c^*} x_{n^*} \neq 0$.
Therefore $\alpha(\pis\ttt{\Nset} n, \pis\ttt{\Mset} m) = \alpha(n, m) \; \forall n,m$, which by definition of $\aut(\Str)$ means $(\pis\ttt{\Nset}, \pis\ttt{\Mset}) \in \aut(\Str)$.
\end{proof}

\begin{proof} of Proposition~\ref{th:dense_design}\\
  To prove $\GG\ttt{\Mset,\Nset} \leq \aut(\Str)$ we simply show that all $(\gg\ttt{\Nset},\gg\ttt{\Mset}) \in \GG\ttt{\Nset, \Mset}$ preserve the relations in $\aut(\Str)$.
  From \cref{eq:aut_bipartite2}, 
  \begin{align*}
    &\gg\ttt{\Nset,\Mset} = (\gg\ttt{\Nset},\gg\ttt{\Mset}) \in \aut(\Str) \Leftarrow\\
    &\bigg( (n,m) \in \Rel_{p,q}  \; \Leftrightarrow\; \ (\gg\ttt{\Nset}n,\gg\ttt{\Mset}m) \in \Rel_{p,q} \bigg)  \quad \forall (p,q), n, m 
  \end{align*}
  The r.h.s holds for all $(\gg\ttt{\Nset},\gg\ttt{\Mset}) \in \GG\ttt{\Nset, \Mset}$ because in constructing relations $\Rel_{p,q}$ in the dense design,
  we used edge-orbits:
  $$(n,m) \in \Rel_{p,q} \Leftrightarrow (\gg\ttt{\Nset} n, \gg\ttt{\Mset} m) \in \Rel_{p,q} \quad \forall (p,q), n, m.$$
  Therefore $\gg\ttt{\Nset,\Mset} \in \GG\ttt{\Mset,\Nset} \Rightarrow \gg\ttt{\Nset,\Mset} \in \aut(\Str)$.
\end{proof}

\begin{proof} of Theorem~\ref{th:design}\\
  We first show that any permutation $(\gg\ttt{\Nset},\gg\ttt{\Mset}) \in \GG\ttt{\Nset,\Mset}$
  is also in $\aut(\Str)$. The major part of the proof is to show that when
  $\GG\ttt{\Nset}$ and $\GG\ttt{\Mset}$ are semi-regular, then $|\aut(\Str)| \leq |\GG\ttt{\Nset,\Mset}|$.
  Combination of these two proves $\aut(\Str)  = \GG\ttt{\Nset,\Mset}$.

  \textbf{I)} to prove that $(\hh\ttt{\Nset},\hh\ttt{\Mset}) \in \GG\ttt{\Nset,\Mset} \Rightarrow (\hh\ttt{\Nset},\hh\ttt{\Mset}) \in \aut(\Str)$,
  we simply apply $(\hh\ttt{\Nset},\hh\ttt{\Mset})$ to an arbitrary edge $(m,n)$ in a relation of $\Str$.
  According to \cref{eq:str_general} $$\Rel_{p,q, \gr{a}} = \{ (\gg\ttt{\Nset} \gr{a} n_p, \gg\ttt{\Mset} m_q) \mid (\gg\ttt{\Nset},\gg\ttt{\Mset}) \in \GG\ttt{\Nset, \Mset} \}.$$
  Application of $(\hh\ttt{\Nset},\hh\ttt{\Mset})$ to $(\gg\ttt{\Nset} \gr{a} n_p, \gg\ttt{\Mset} m_q)$ gives
  $(\hh \gg\ttt{\Nset} \gr{a} n_p, \hh \gg\ttt{\Mset} m_q) = (\gg'\ttt{\Nset} \gr{a} n_p, \gg'\ttt{\Mset} m_q) \in \Rel_{p,q,\gr{a}}$.
  From \cref{eq:aut_bipartite2}, it follows that $\GG\ttt{\Nset,\Mset} \leq \aut(\Str)$.


\textbf{II)}
  For this part, we use the orbit-stabilizer theorem. The orbit of each pair $(n,m) \in \Rel_{p,q,a}$
wrt $\HH\ttt{\Nset, \Mset}$ is defined as $\HH\ttt{\Nset,\Mset} (n,m) = \{ (\hh\ttt{\Nset} n,\hh\ttt{\Mset} m) \mid \hh\ttt{\Nset,\Mset} \in \HH\ttt{\Nset,\Mset}\}$.
The \textbf{stabilizer} $\HH\ttt{\Nset,\Mset}^{(n,m)}$ of $(n,m) \in \Rel_{p,q,a}$ is
$\HH\ttt{\Nset,\Mset}^{(n,m)} = \{\hh\ttt{\Nset, \Mset} \in \HH\ttt{\Nset, \Mset} \mid \hh\ttt{\Nset, \Mset} (n,m) = (n,m)\}$,
the group of all actions that fix $(n,m)$. 
The orbit-stabilizer theorem states that $|\HH\ttt{\Nset,\Mset}| = |\HH\ttt{\Nset, \Mset}^{(n,m)}| \times |\HH\ttt{\Nset,\Mset} (n,m)|$.
In our argument, we apply this theorem to bound $|\aut(\Str)|$ using $|\aut(\Str)^{(n,m)}|$
and $|\aut(\Str) {(n,m)}|$. 

The orbit-size, $|\aut(\Str) {(n,m)}|$, for a pair $(n,m)$ is bounded by the size of its relation $|\Rel_{p,q,\gr{a}}|$, 
for some $p,q,\gr{a}$. This is because, according to \cref{eq:aut_bipartite}, 
$$\pis \in \aut(\Str) \Rightarrow ((n,m) \in \Rel_{p,q,\gr{a}}  \Rightarrow\; \pis (n,m) \in \Rel_{p,q,\gr{a}}).$$
From \cref{eq:str_general}, $|\Rel_{p,q,a}| = |\GG\ttt{\Nset,\Mset}|$, and therefore $|\aut(\Str) {(n,m)}| < |\GG\ttt{\Nset, \Mset}|$.

Now, it only remains to show that if $\GG\ttt{\Nset}$ and $\GG\ttt{\Mset}$ are regular orbits (or semi-regular), 
the stabilizer is trivial $\aut(\Str)^{(n,m)} = \{\gr{e}\}$. Because in this case the size of $\aut(\Str)$
is bounded by the size of orbit $|\aut(\Str)| = |\aut(\Str)(n,m)| \leq |\GG\ttt{\Nset, \Mset}|$, which combined with the result of part {(I)} gives $\GG\ttt{\Nset,\Mset} = \aut(\Str)$.

Since, according to our assumption, $\GG\ttt{\Nset}$ acts regularly on $\GG\ttt{\Nset} n_p \; \forall p$,
going back to definition of $\Rel_{p,q,\aa} = \{(\gg\ttt{\Nset} \aa n_p, \gg\ttt{\Mset} m_q) \mid \gg\ttt{\Nset, \Mset} \in \GG\ttt{\Nset, \Mset}\}$,
this (see definition of regularity) implies that for each $n \in \GG\ttt{\Nset} n_p$, $\aa \in \gr{A}$ and $m_q$, we can identify a single $\gg'\ttt{\Nset} \in \GG\ttt{\Nset}$
such that for some $(n, m) = (\aa \gg'\ttt{\Nset} n_p, \gg'\ttt{\Mset} m_q) \in \Rel_{p,q,\aa}$. This means that \textit{the edges (or pairs) adjacent to each node $n \in \GG\ttt{\Nset} n_p$
all have distinct colors}. The same argument using regularity of $\GG\ttt{\Mset}$-action on $\GG\ttt{\Mset} m_q \; \forall q$ shows that
edges (or pairs) adjacent to $m \in \GG\ttt{\Mset} m_q$ all have distinct colors.

Therefore if we fix a pair $(m,n)$, all their neighboring edges (adjacent on $n$ or $m$) are unambiguously fixed. The same goes for the neighbors of the newly fixed nodes and so on.
If we can show that the bipartite graph representing $\Str$ is \textit{connected}
then fixing a pair guarantees that all pairs in all relations of $\Str$ are fixed and therefore
$(n,m)$ has a trivial stabilizer.

Two properties guarantee the connectedness of $\Str$:
\begin{itemize}
\item Since $\gr{A} = \gr{A}^{-1}$ is a generating set of $\GG$, the bipartite subset consisting of subset of nodes $\GG\ttt{\Nset} n_p$ and $\GG\ttt{\Mset} m_q$ are connected.
  To show this, it is enough to show that we can reach any node $n_z$ starting from an arbitrary representative $n_p$ and zigzagging through the bipartite structure.
  Since $n_z,n_p \in \GG\ttt{\Nset} n_p \Rightarrow \exists \gg_z \in \GG\ttt{\Nset}\; \text{s.t.} \; n_z = \gg_z n_p$. Since $<\gr{A}> = \GG\ttt{\Nset, \Mset}$, we can write
    $\gg_z = \aa_1 \ldots \aa_L$.
The path that starts from $n_p$ and takes the connections corresponding to $\Rel_{p,q,\aa_L}, \Rel_{p,q,\aa^{-1}_{L-1}}, \Rel_{p,q,\aa_{L-2}},\ldots, \Rel_{p,q,\aa^{-1}_{1}}$ takes us through a zigzag path from $n_p$ to $n_z$. 
\item Since we have a relation $\Rel_{p,q, \aa}$ for all pairs $p,q$, all the induced bipartite subgraphs on $\GG\ttt{\Nset} n_p$-$\GG\ttt{\Mset} m_q$ are connected.
\end{itemize}
This proves that the whole bipartite
graph is connected and unambiguously fixed if we fix any pair $(n,m)$. Therefore, $(n,m)$ has a trivial stabilizer, proving that $\aut(\Str) = \GG\ttt{\Nset, \Mset}$.
\end{proof}

\begin{proof} of Corollary~\ref{cor:str_general}
  Follows directly from Theorems~\ref{th:main}~and~\ref{th:design}.
\end{proof}


  \begin{proof} of Claim~\ref{claim:group_conv}\\
To see this, note that $\GG\ttt{\Mset}$ acts on $\Mset = \GG$ regularly, with the natural (group) action $\gg \hh$.
Set the representative from the resulting single orbit as $m_q = \gr{e}$.
Then \cref{eq:layer_str} becomes
$\phis = [\phi_\gg]_{\gg \in \GG}$ with components
\begin{align}
  \label{eq:layer_gconv}
  \phi_\gg(\xx; \thetas) = \sigma \bigg (\sum_{1 \leq p \leq P}\sum_{\aa \in \gr{A}} \theta_{\aa, p} x_{ \gg\ttt{\Nset} \aa n_p} \bigg )
\end{align}
If we further \textit{tie the parameters across the orbits} so that $\theta_{\aa, p} = \theta_{\aa, p'} \forall p,p'$,
the \cref{eq:layer_gconv} above is equivalent to formulation of \citep{cohen2016group} for a single input/output channels (see \cref{sec:multiple_channels} for multiple channels).    
  \end{proof}

\begin{proof} of Corollary~\ref{cor:graph_conv}\\
First we show this assuming a single channel $K=1$. For multiple channels see \cref{sec:multiple_channels}.

Consider the bipartite structure constructed from $\Lambda$:
$\Str = (\Nset, \Nset, \{\{(n,n) \mid n \in \Nset \}, \{(n,n') \mid (n,n') \in \EE(\Lambda)\} \})$.
Applying the result of Theorem~\ref{th:main} using $\sigmas (\xx) = \xx$ tells us that the function $\mathbf{A}\xx$ is uniquely $\aut(\Str)$-equivariant -- that is
$\pis (\mathbf{B}\xx_{\cdot,k}) = \mathbf{B} (\pis\xx_{\cdot,k}) \forall \pi \in \aut(\Str)$.
Because of the relation $\Rel_{c^*} = \{\{(n,n) \mid n \in \Nset \}$ in $\Str$, the same bipartite structure $\Str$,
can be interpreted as a digraph; here with a single color, since $\Str$ has only one relation in addition to $\Rel_{c^*}$.
Since this relation defines $\Lambda$,  $\aut(\Str) = \aut(\Lambda)$, which means  $\mathbf{B}\xx$ is uniquely $\aut(\Lambda)$-equivariant.
\end{proof}

\section{Background on Permutation Groups}\label{app:background}
Let $\xx = [x_1,\ldots,x_N] \in \XX^{N}$ be a vector of $N$ variables taking value in the same domain $\XX$.
A \textit{group} $\GG$ is a set, equipped with a binary operation, with the following properties: \textbf{I}) $\GG$ is closed under its binary operation; \textbf{II}) the group operation is associative --\ie $(\gg_1 \gg_2)  \gg_3 = \gg_1 (\gg_2  \gg_3) \forall \, \gg_1,\gg_2,\gg_3 \in \GG$; \textbf{III}) there exists an identity $\gr{e} \in \GG$ such that $\gg \gr{e} = \gr{e} \gg = \gg$ and ; \textbf{IV}) every element $\gg \in \GG$ has an inverse $\gg^{-1} \in \GG$, such that $\gg  \gg^{-1} = \gg^{-1} \gg = \gr{e}$. A subset $\HH \subseteq \GG$ is a subgroup of $\GG$ ($\GG \leq \HH$) iff $\HH$ equipped with the binary operation of $\GG$ forms a group. Moreover, if $\HH$ is a proper subset of $\GG$, $\HH$ is a proper subgroup of $\GG$, $\HH < \GG$.  
Two groups are isomorphic $\GG \cong \HH $ if there exists a bijection $\beta: \GG \to \HH$, such that $\gg_1 \gg_2 = \gg_3 \Leftrightarrow \beta(\gg_1) \beta(\gg_2) = \beta(\gg_3) \forall \gg_1,\gg_2, \gg_3$. If this last relation holds for a surjective mapping (not necessarily one-to-one) then $\beta$ is a homomorphic mapping and $\HH$ is isomorphic to a subgroup of $\GG$.

\textbf{Cayley Diagram.} 
The set $\gr{A} \subseteq \GG$ is called the \textbf{generating set} of $\GG$ ($<\gr{A}> = \GG$), 
iff every member of $\GG$ can be expressed as a combination of members of $\gr{A}$. 
If the generating set is closed under inverse $\aa \in \gr{A} \Rightarrow \aa^{-1} \in \gr{A}$ we call it a \textit{symmetric generating set}.
$\gr{A}$ is the \textit{minimal} generating set if it has the least number of members among the
generating sets of $\GG$. Note that the minimal generating sets are generally not unique.
The size of the minimal generating set of a group $\GG$ becomes important because, 
the number of parameters in our parameter-sharing scheme grows linearly with $|\gr{A}|$.
A group $\GG$ is often visualized by its Cayley diagram; a colored digraph in which the node-set is $\GG$ and 
directed edge $(\gg, \aa \gg) \forall \gg \in \GG, \aa \in \gr{A}$ is colored by $\aa \in \gr{A}$. \cref{fig:main}(lower-left) shows the
Cayley diagram of $\GG = \gr{D}_5$.

\subsection{Discrete Group Action}
We are interested on the way a group ``acts'' on the input and output of a deep network.
Function $\gamma: \GG \times \XX \to \XX$ is the left \textit{action} of group $\GG$ on $\xx$ iff
\textbf{I}) $\gamma(\gr{e}, \xx) = \xx$ and; \textbf{II}) $\gamma(\gr{g}_1, \gamma(\gr{g}_2, \xx)) = \gamma(\gr{g}_1 \gr{g}_2, \xx)$.\footnote{All the following definitions and results may be extended to the ``right'' group action by substituting $\gg \leftrightarrow \gg^{-1} \forall \gg \in \GG$.}

For our purpose we limit this action to actions on the indices $\Nset = \NN$ of $\xx = [x_n]$ -- \ie
function $\gamma: \GG \times \Nset \to \Nset$ satisfies $\gamma(\gr{e}, n) = n$ and $\gamma(\gr{g}_1, \gamma(\gr{g}_2, n)) = \gamma(\gr{g}_1  \gr{g}_2, n)$.
We often use $\gg n$ as a shorthand for $\gamma(\gg, n)$, and also use $\gg \Nset$ to denote $\{\gg n \mid n \in \Nset\}$.
The action of $\gg$ on a vector/sequence $\Nvec = \nn$ is defined similarly $\gg \Nvec \defeq [\gg 1,\ldots, \gg N]$.
Considering this, the $\GG$-action on $\xx = [x_1,\ldots,x_N]$ is $\gg \xx \defeq [x_{\gg 1}, \ldots x_{\gg N}]$.

From the properties of group and its action it follows that
  $\gamma(\gg, \cdot): \Nset \to \Nset$ is a bijection with $\gamma^{-1}(\gg, n) = \gamma(\gg^{-1}, n) \, \forall n \in \Nset, \gg \in \GG$.
Since $\Nset$ is a finite set, this bijection for each $\gg \in \GG$ is a permutation of $\Nvec$
 -- \ie $\gg \Nvec \defeq [\gamma(\gg, 1), \ldots, \gamma(\gg, N)]$ is a permutation of $\Nvec$. 
Let $\GG\ttt{\Nset} = \{\gamma(\gg, \cdot) \mid \gg \in \GG\}$ with (function composition as the binary group operation) 
denote the group of permutations of $\Nvec$ induced by $\gg \in \GG$.
This group is a subgroup of the \textit{symmetric group} $\gr{S}\ttt{\Nset}$; the group of all $N!$ permutations of $\Nvec$.
$\GG\ttt{\Nset}$ captures the structure of $\GG$ when it acts on the set $\Nset$
and it is indeed a homomorphic image of $\GG$.
We use $\gg\ttt{\Nset}$ to denote  $\gamma(\gg, \cdot)$, the the image of $\gg\in\GG$ in $\GG\ttt{\Nset}$. 



\subsubsection{Properties of Group Action}
$\GG$-action is \textit{faithful} iff two groups are isomorphic $\GG \cong \GG\ttt{\Nset}$.
In this case all actions of $\gg \in \GG$ are distinct permutations
-- that is $\gg \Nvec \neq \gg' \Nvec \forall \gg, \gg' \in \GG$.
Given any $\GG$-action on $\Nset$ we can obtain its faithful subgroup that is isomorphic to $\GG\ttt{\Nset}$.
The importance of faithfulness of $\GG$-action is because it preserves the structure of $\GG$,
and if an action is not faithful, we might as well focus on $\GG\ttt{\Nset}$-action.

Given any unfaithful  $\GG$-action $\gamma: \GG \times
\Nset \to \Nset$, let $\gr{K}_{\gamma}$ be the normal subgroup of $\GG$ that corresponds to identity permutation
--\ie $\gr{K}_{\gamma} = \{ \gg \in \GG\; \mid\; \gamma(\gr{g}, n) = n \forall n \in \Nset\}$.
One obtains the group $\GG\ttt{\Nset}$ that acts \textit{faithfully} on $\Nset$ as the \textit{quotient group} $\GG\ttt{\Nset} = \GG / \gr{K}_{\gamma}$.

We now define some group properties that are important in guaranteeing the ``strict'' equivariance with respect to $\GG$-action.
$\GG$-action on $\Nset$ is \textit{transitive} iff $\forall n_1,n_2 \in \Nset$, there exists at least one action $\gg \in \GG$ such that $\gg n_1 = n_2$. 
The group action is free or \textit{semi-regular} iff $\forall n_1,n_2 \in \Nset$, there is at most one $\gg \in \GG$ such at $\gg n_1 = n_2$, and the action is \textit{regular} iff it is both transitive and free -- \ie for any pair $n_1,n_2 \in \Nset$, there is uniquely
one $\gg \in \GG$ such that $\gg n_1 = n_2$. Any free action is also faithful. 

\subsubsection{Orbits}
Given $\GG$-action on $\Nset$, the \textit{orbit} of  $n \in \Nset$ is all the members to which it can be moved, 
$\GG n = \{\gg n \mid n \in \Nset\}$. The orbits of 
$n \in \Nset$ form an equivalence relation, where $n \sim n' \Leftrightarrow \exists g \;\text{s.t.,}\; n = \gg n' \Leftrightarrow n \in \GG n' \Leftrightarrow n' \in \GG n$.
This equivalence relation \textit{partitions} $\Nset$ into orbits $\Nset = \bigcup_{1 \leq p \leq P}  \GG n_p$, where $n_p$ is an arbitrary representative of the partition 
$\GG n_p \subseteq \Nset$.
Note that the $\GG$-action on $\Nset$ is always transitive on its orbits -- that is for any $n,n' \in \GG n_p$, there is at least one $\gg \in \GG$ such that $n = \gg n'$. Therefore, for a semi-regular $\GG$-action, the action of $\GG$ on the orbits $ \GG n_p \forall 1 \leq p \leq P$ is regular.
As we see the number of distinct parameters in our parameter-sharing scheme grows with the number of orbits.


\textbf{Cycle Notation.} To explicitly show the action of $\gg \in \GG$ on the set
$\Nset$, we sometimes use the {cycle notation} of a permutation.
Any permutation $\pis \in \gr{S}\ttt{\Nset}$ is decomposable to product of disjoint cycles. 
A cycle of length $d$, $(b_1,\ldots,b_d)$ sends $b_i \to b_{i+1 \bmod d}$. Here $b_i \in \NN$ and a cycle acts on a subset of $\Nset$. 
For example, the action of $(1,3,2)$ on $[1,\ldots,6]$ is 
$[3,1,2,4,5,6]$. We can write the permutation $\gg$ where $ \gg [1,\ldots,6]= [3,1,2,5,4,6]$ as the {product of disjoint cycles} 
$\{(1,3,2),(6),(4,5)\} = \{(1,3,2),(4,5)\}$.

}
\end{document}